\documentclass{article}
\usepackage{spconf,amsmath,graphicx}

\usepackage{enumitem}
\setlist{nosep, leftmargin=14pt}

\usepackage{mwe} 
\usepackage{multirow}
\usepackage{booktabs}
\usepackage{url}

\title{Uncertainty-Aware Structured Data Extraction from\\ Full CMR Reports via Distilled LLMs}

\name{\hspace{-2pt}Yi Yu$^{\dagger}$\hspace{-2pt}, Parker Martin$^{\dagger}$\hspace{-2pt}, Zhenyu Bu$^{\dagger}$\hspace{-2pt}, Yixuan Liu$^{\dagger}$\hspace{-2pt}, Yi-Yu Zheng$^{\ddagger}$\hspace{-2pt}, Orlando Simonetti$^{\dagger}$\hspace{-2pt}, Yuchi Han$^{\dagger}$\hspace{-2pt}, Yuan Xue$^{\dagger}$\vspace{-5pt}}
\address{$^{\dagger}$The Ohio State University \quad $^{\ddagger}$NetMind.AI\vspace{-5pt}}
\begin{document}
%
\maketitle
\begin{abstract}

Converting free-text cardiac magnetic resonance (CMR) reports into auditable structured data remains a bottleneck for cohort assembly, longitudinal curation, and clinical decision support. We present CMR-EXTR, a lightweight framework that converts free-text CMR reports into structured data and assigns per-field confidence for quality control. A teacher–student distillation pipeline enables fully offline inference while limiting manual annotation. Uncertainty integrates three complementary principles—distribution plausibility, sampling stability, and cross-field consistency—to triage human review. Experiments show that CMR-EXTR achieves 99.65\% variable-level accuracy, demonstrating both reliable extraction and informative confidence scores. To our knowledge, this is the first CMR-specific extraction system with integrated confidence estimation. The code is available at \url{https://github.com/yuyi1005/CMR-EXTR}.
\end{abstract}
%
%
\vspace{-1pt}
\section{Introduction}
\label{sec:intro}
\vspace{-1pt}

\begin{figure*}[t]
\includegraphics[width=\linewidth]{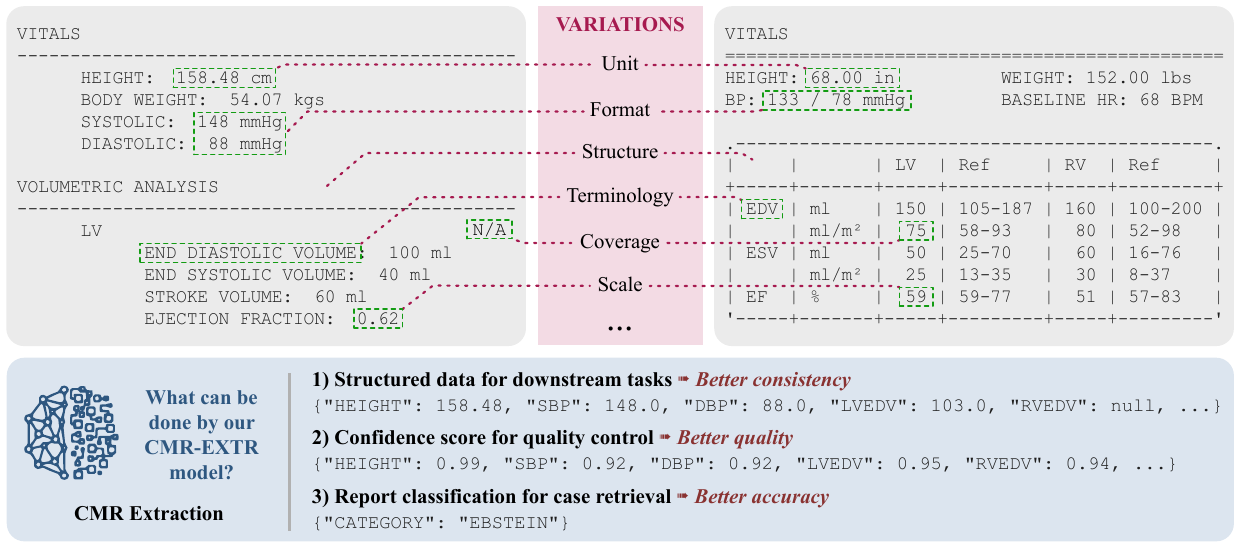}
\vspace{-18pt}
\caption{Motivation and challenges of our work. The upper panel compares two reports, highlighting variations in report structure and content. The lower panel illustrates the capabilities of our proposed model.}
\label{fig:motivation}
\vspace{-6pt}
\end{figure*}

Cardiac magnetic resonance (CMR) is a reference standard for biventricular function, chamber size, and tissue characterization, and its free-text reports remain the authoritative record of quantitative measurements and diagnostic impressions. Turning these reports into structured data enables scalable cohort assembly, longitudinal research curation, and integration with clinical decision support. However, heterogeneity in terminology, units, formatting, and completeness (Fig.~\ref{fig:motivation}) makes end-to-end, fully automated parsing difficult. Manual curation is also costly, slow, and raises privacy concerns. These realities motivate methods that are accurate, auditable, and deployable under offline constraints typical for protected health information~\cite{wang2024screening}. 

Large language models (LLMs) now achieve strong performance across diverse language tasks, exemplified by GPT-4~\cite{openai2024gpt4}, and a rapidly maturing ecosystem of open-weight models broadens accessibility. Recent families such as GPT-OSS~\cite{openai2025gptoss} and Llama~\cite{grattafiori2024llama3} are particularly attractive for report processing in privacy-sensitive settings because they support local, offline inference without transmitting data externally. 

Building on these advances, a growing literature investigates clinical information extraction from free-text electronic health records (EHRs). Systematic and scoping reviews highlight the promise of LLMs and domain-tuned models for structured data capture~\cite{wieland2024natural, nunes2024health}. Concrete exemplars include a two-stage pipeline that first converts reports to tables and then applies an LLM agent for quality review~\cite{bisercic2023interpretable}, and a fine-tuned Llama-2-7B model that extracts donation-relevant measurements from annotated medical notes~\cite{adam2024clinical}. Within cardiovascular applications, sentence-level aortic measurements have been extracted from CT reports with instruction-tuned models, with Llama-3.1-8B performing best among baselines~\cite{erez2025instruction} compared to BERT-style encoders~\cite{huang2019clinicalbert}; BERT-based pipelines have also been trained on CMR reports to extract targeted quantitative values~\cite{singh2022one}. Beyond extraction, open-source LLMs have been benchmarked for CMR report classification~\cite{amirrajab2025comparative}, and synthetic-data strategies have been explored to improve classification efficiency in our prior work~\cite{martin2025efficient}. 

Despite this progress, three practical gaps constrain CMR report curation at scale. First, many systems operate at the \emph{sentence} rather than \emph{report} level, requiring bespoke segmentation and limiting their ability to reconcile cross-sentence dependencies and units~\cite{erez2025instruction, martin2025efficient}. Second, most pipelines rely heavily on manual annotation to create training corpora~\cite{bisercic2023interpretable, adam2024clinical, singh2022one}, which impedes scalability. Third, and most critical for safe use, prior work generally lacks \emph{per-field} uncertainty estimates that guide human review, surface inter-value inconsistencies (e.g., between stroke volume and ejection fraction), and help identify potential issues in either the extraction or the original report. As a result, quality evaluation often reduces to ad hoc spot checks rather than principled triage.

In light of these limitations, we propose \text{CMR-EXTR}, a compact, CMR-specialized LLM that parses the \emph{entire} report in a single pass and outputs a standardized 52-field JSON spanning vitals, biventricular volumes and function (absolute and indexed), chamber dimensions, atrial metrics, and tissue characterization. CMR-EXTR distills a 20B GPT-OSS \cite{openai2025gptoss} teacher into a 1B student for efficient offline use, and couples extraction with per-field uncertainty that blends three complementary signals: \emph{distribution} plausibility under reference ranges, \emph{stability} across controlled stochastic generations, and \emph{consistency} with physiologic/algebraic constraints using 22 formulas; low-confidence fields are escalated for review. The same model also outputs a report-level diagnostic category to support rapid case retrieval.

On 1,100 CMR reports from OSU hospital (20\% held out for testing), CMR-EXTR significantly outperforms both the SOTA zero-shot model and the undistilled 1B baseline. Ablations confirm that: concise field descriptions and restricting the loss to the JSON target improve fidelity, and the uncertainty score is strongly discriminative for triage (42\% error when confidence $<0.7$ vs.\ 1\% when $>0.7$). In summary, our contributions are threefold: \textbf{1)} a full-report, offline CMR extraction framework that produces a 52-field JSON in one pass via efficient distillation; \textbf{2)} a per-field uncertainty scheme integrating distribution, stability, and multi-formula consistency checks to surface potential inconsistencies for quality control; and \textbf{3)} a practical, privacy-compatible pipeline that minimizes annotation while delivering state-of-the-art extraction and strong classification performance.

\vspace{-1pt}
\section{Methodology}
\label{sec:method}
\vspace{-1pt}

\begin{figure*}[t]
\includegraphics[width=\linewidth]{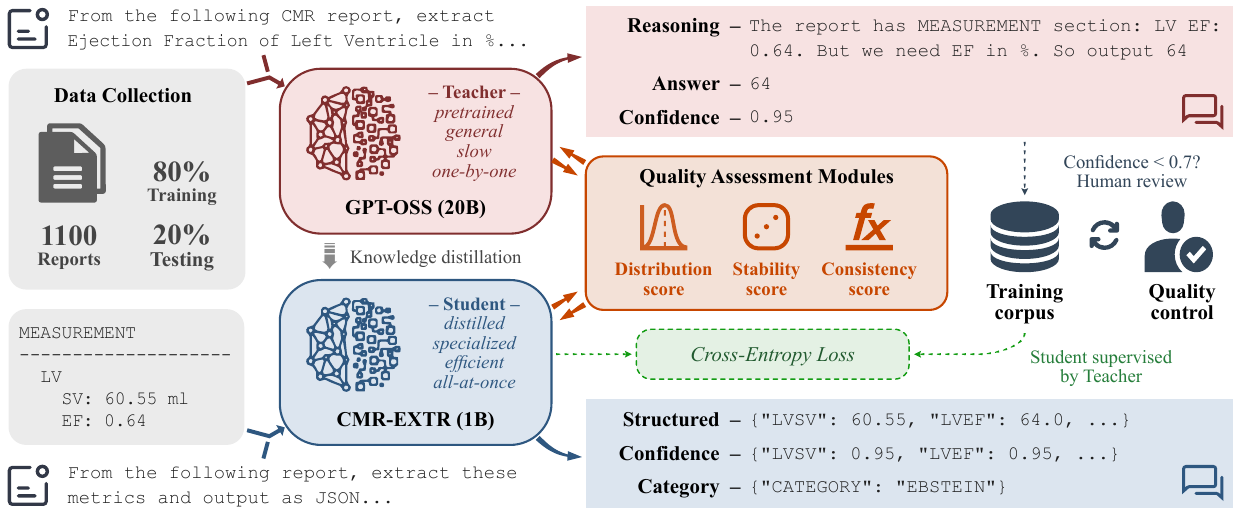}
\vspace{-18pt}
\caption{The overall framework of our method to distill knowledge from teacher model to student model. We devise and three quality assessment modules (see Sec.~\ref{sec:method}), incorporated into both models, significantly alleviating the burden of human review.}
\label{fig:framework}
\vspace{-6pt}
\end{figure*}

The overall framework of our work, illustrated in Fig.~\ref{fig:framework}, follows a knowledge distillation paradigm with human-in-the-loop quality control.
Different from existing work~\cite{bisercic2023interpretable, adam2024clinical, erez2025instruction, singh2022one} directly using manual annotation, we utilize the zero-shot ability of a powerful teacher model to provide the knowledge. During this process, 52 values are extracted one by one from each report to establish the training corpus. 
Subsequently, values with confidence scores below 0.7 are manually reviewed to ensure corpus quality. 
This human-in-the-loop quality control strategy produces a cleaner training corpus with minimal annotation effort.
Finally, the reviewed training corpus is used to train the student model.

\vspace{2pt}\noindent\textbf{Model selection.} Based on the above knowledge distillation concept, the teacher model is allowed to be large and slow, but with powerful pretrained ability. The recent open-weight model, GPT-OSS-20B~\cite{openai2025gptoss}, is then chosen. Comparably, the student model needs to be smaller and efficient, so we base our CMR-EXTR model on Llama-3.2-1B~\cite{grattafiori2024llama3}.

\vspace{2pt}\noindent\textbf{Values extracted.} We extract 52 variables from each report, and the output will be ``null" if it is not present in the report.
\textbf{1) Vital signs.}
Height, Weight, Body Surface Area (BSA), Systolic/Diastolic Blood Pressure (SBP, DBP), Baseline Heart Rate (BHR).
\textbf{2) Ventricular function.} 
Left Ventricle End-Diastolic Volume (LVEDV), End-Systolic Volume (LVESV), Cardiac Output (LVCO), Total Mass (LVMASS), Stroke Volume (LVSV), Ejection Fraction (LVEF), corresponding right ventricular parameters, and their BSA-indexed counterparts.
\textbf{3) Ventricular structure.}
Left/Right Ventricular End-Diastolic/-Systolic Diameter (LVEDD, RVEDD, LVESD, RVESD), Anteroseptal/Inferolateral Wall Thickness (LVAWT, LVIWT).
\textbf{4) Atrial structure.}
Left Atrial Volume (LAV) and indexed values (LAVI), Left Atrial Area/Length in two-/four-chamber views (LAA2CH, LAA4CH, LAL2CH, LAL4CH), and corresponding right atrial parameters.
\textbf{5) Tissue characteristics.} 
Hematocrit (HCT), Pre-/Post-contrast T1 of Myocardium/Blood (PRET1M, PRET1B, POSTT1M, POSTT1B), Extracellular Volume Fraction (ECV).

\vspace{-6pt}
\subsection{Data Preparation}\label{sec:method-data}
\vspace{-2pt}

We collect 1100 CMR reports for this study from OSU hospital. No manual annotation is involved during data preparation. Instead, we review fewer than 1\% of the values produced by the teacher model with low confidence scores identified by our quality assessment module, which substantially reduces the labor-intensive annotation effort. The reports are exported with a disease categories, allowing us to train and evaluate a classification task. The categories include Coronary Artery Disease (CAD), Hypertrophic Cardiomyopathy (HCM), Dilated Cardiomyopathy (DCM), Ebstein's Anomaly (Ebstein), and Pulmonary Arterial Hypertension (PAH).

\vspace{2pt}\noindent\textbf{Privacy control.} All the models used in this studies are deployed offline. While our model is not designed to learn the private information from training corpus, to avoid any misuse of our open-weight model, we remove lines that contain the following keywords from all reports: MRN, Name, DOB, Date, Account, Physician, Nurse, Technologist.

\vspace{-6pt}
\subsection{Quality Assessment Modules}
\vspace{-2pt}

To equip the model with the capability of estimating confidence scores for each output value, we introduce three principles grounded in distribution, stability, and consistency. The final score is the average of the three scores.

\vspace{2pt}\noindent\textbf{Distribution score.}
This module assesses whether the extracted values fall within a reasonable range. For each target variable, we compute a score based on how likely the value is under its empirical normal distribution, defined by the mean ($\mu$) and standard deviation ($\sigma$) obtained from reference~\cite{kawelboehm2025society}:
\begin{equation}
\setlength{\abovedisplayskip}{6pt}
\setlength{\belowdisplayskip}{6pt}
\scalebox{1}{$S_\text{dist} = \exp\left(-\frac{1}{2}\left(\frac{v - \mu}{\alpha\sigma}\right)^2\right)$}
\end{equation}
where $v$ denotes the extracted value, $\alpha$ is set to 6 by default.

Since gender can affect the distribution and most CMR reports lack explicit gender information, we calculate two scores based on both male and female distributions and take the higher one. If no reference distribution is available for a specific variable, the score defaults to 0.7.

\begin{table*}[t]
\small
\setlength{\tabcolsep}{2.5mm}
\centering
\caption{Quantitative results for CMR report extraction.}
\vspace{6pt}
\begin{tabular}{c|c|c|c|c|c|c|c}
\toprule
\multirow{2}{*}{\textbf{Model}} & \multicolumn{2}{c|}{\textbf{Accuracy}} & \multicolumn{5}{c}{\textbf{Error Breakdown}}   \\
 & Variable-Level & Report-Level & Omission & \,Inexact\, & Confusion & ~Invalid~ & ~~Total~~ \\
\midrule
GPT-OSS-20B {\scriptsize(FREE-TEXT)} & 90.78\% & 46.82\% & 166 & 5 & 104 & 780 & 1055 \\
GPT-OSS-20B {\scriptsize(STRUCTURED)} & 96.07\% & 49.09\% & 102 & 6 & 186 & 156 & 450 \\
Llama-3.2-1B {\scriptsize(FREE-TEXT)} & 25.31\% & 0.00\% & 2519 & 590 & 1432 & 4004 & 8545 \\
Llama-3.2-1B {\scriptsize(STRUCTURED)} & 47.91\% & 3.18\% & 162 & 163 & 5634 & 0 & 5959 \\
CMR-EXTR (ours) & \textbf{99.65\%} & \textbf{89.55\%} & \textbf{15} & \textbf{5} & \textbf{20} & \textbf{0} & \textbf{40} \\
\bottomrule
\end{tabular}
\label{tab:main}
\vspace{-6pt}
\end{table*}

\begin{table}[t]
\small
\setlength{\tabcolsep}{1.95mm}
\centering
\vspace{-10pt}
\caption{Results for CMR report classification.}
\vspace{6pt}
\begin{tabular}{c|c|c|c|c}
\toprule
\textbf{Model} & \textbf{Acc.} & \textbf{Prec.} & \textbf{Rec.} & \textbf{F1} \\
\midrule
GPT-OSS-20B & 91.43\% & 90.70\% & 93.78\% & 91.96\% \\
Llama-3.2-1B & 69.95\% & 61.60\% & 60.53\% & 59.17\% \\
CMR-EXTR (ours) & \textbf{97.04\%} & \textbf{98.42\%} & \textbf{96.19\%} & \textbf{97.24\%} \\
\bottomrule
\end{tabular}
\label{tab:cls}
\vspace{-6pt}
\end{table}

\vspace{2pt}\noindent\textbf{Stability score.}
This module evaluates the stability of the extracted values when slight randomness is introduced into generation. LLMs typically employ token sampling, controlled by a temperature between 0 and 1, to adjust the degree of randomness. We run the model three times a temperature of 0.3, obtaining three outputs $v_1, v_2, v_3$. A vote-based mechanism is used to determine the final value.
Meanwhile, we compute the stability score between $v_i$ and $v_j$ as:
\begin{equation}
\setlength{\abovedisplayskip}{6pt}
\setlength{\belowdisplayskip}{6pt}
\scalebox{1.0}{$s_\text{i,j} = \exp\left(-\beta \cdot \frac{(v_i - v_j)}{v_i + v_j}\right)$}
\label{equ:scoreij}
\end{equation}
where $\beta$ is a scaling factor set to 2 by default. Specially, $s_\text{i,j}$ is set to 1 when both $v_i$ and $v_j$ are ``null", and set to 0.5 when one is ``null" and the other is not. The final stability score is calculated as $S_\text{stab} = ({}s_\text{1,2} + s_\text{1,3} + s_\text{2,3}) \,/\, 3$, where a higher score indicates that the model produces stable results under small random perturbations, reflecting higher reliability.

\vspace{2pt}\noindent\textbf{Consistency score.} This score evaluates the relationships between different values in a CMR report. Certain variables are mathematically related. For example, we have
$\text{LVEF} = \text{LVSV} \,/\, \text{LVEDV} \times 100$.
We calculate the right-hand side of the formula and then apply Eq.~\ref{equ:scoreij} to obtain the consistency score $S_\text{cons}$. In total, we employ 22 formulas (refer to our code). When a variable appears in multiple formulas, we take the average scores across all formulas. When a variable is not involved in any formula, the score is set to 0.7 by default.

\vspace{-6pt}
\subsection{Report Classification}
\vspace{-2pt}

We also add a ``CATEGORY" entry at the end of each JSON when generating the training corpus. The labels are derived from the ground truth during data preparation (see Sec.~\ref{sec:method-data}). During inference, in addition to extracting values from the report, the LLM also predicts the corresponding category.

\vspace{-1pt}
\section{Experiments}
\label{sec:experiments}
\vspace{-1pt}


\noindent\textbf{Experimental setup.} The model is trained on an NVIDIA A100 GPU, using Llama-3.2-1B as the initialization for our CMR-EXTR. We apply LoRA \cite{hu2022lora} fine-tuning with a rank of 16, scaling factor of 32, and dropout rate of 0.05, targeting the query and value projection layers. Training is conducted for 5 epochs with a per-device batch size of 2, gradient accumulation of 1, a learning rate of $2 \times 10^{-4}$, and mixed-precision (FP16) optimization.


\vspace{2pt}\noindent\textbf{Quantitative extraction.} For the extraction, we evaluate the accuracy at both the variable and report levels. Our model achieves a variable-level accuracy of 99.65\%. Each CMR report contains 52 variables and is considered correct only when all are accurately extracted. Under this stricter criterion, our report-level accuracy reaches 89.55\%. For comparison, we also configure GPT-OSS-20B and Llama-3.2-1B to directly output the whole JSON (see Table~\ref{tab:main}). The suffix ``{\scriptsize FREE-TEXT}" denotes unconstrained generation, whereas ``{\scriptsize STRUCTURED}" enforces JSON schema during sampling. Notably, even if set to ``{\scriptsize STRUCTURED}", GPT-OSS-20B sometimes loops in reasoning until all tokens are consumed, failing to produce valid JSON. The results highlight the performance of our model.

To better understand the errors, we further categorize errors into four types (see Table~\ref{tab:main}):
\textbf{1) Omission.} The value is present in the report but missing in the structured output.
\textbf{2) Inexact.} The extracted value differs slightly (within 10\%) from the ground truth.
\textbf{3) Confusion.} The value is incorrectly taken from another field.
\textbf{4) Invalid.} The output fails to follow the required format and cannot be parsed as JSON.


\begin{figure}[t]
\includegraphics[width=\linewidth]{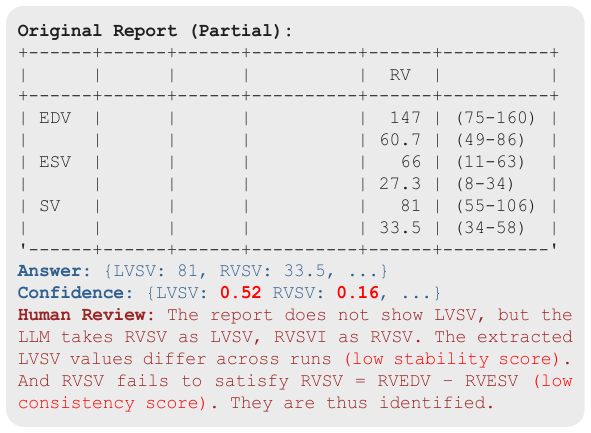}
\vspace{-18pt}
\caption{Exemplary case detected by our quality control.}
\label{fig:example}
\vspace{-6pt}
\end{figure}

\vspace{2pt}\noindent\textbf{Disease classification.} Our model also surpasses the baselines (configured to ``{\scriptsize STRUCTURED}"), achieving 97.04\% accuracy across five disease categories (see Table~\ref{tab:cls}). For comparison, our previous study \cite{martin2025efficient} reports 95\% accuracy on four categories, confirming that our LLM-based design provides an effective solution for both extraction and classification.


\vspace{2pt}\noindent\textbf{Ablation studies.} These experiments are carried out to validate our design: 
\textbf{1) Training detail.} 
First, including descriptions of the 52 variables in the input prompt proves essential, since removing them reduces accuracy (99.65\% vs. 90.31\%). Second, computing the loss only on the JSON portion (by masking other text) improves accuracy (99.65\% vs. 98.40\%).
\textbf{2) Quality control.}
We sample 100 variables with confidence scores below 0.7 and find an error rate of 42\%. An example is shown in Fig.~\ref{fig:example}. In contrast, among 100 randomly selected variables with confidence scores above 0.7, the error rate is only 1\%. This clear discrepancy validates the effectiveness of the proposed quality assessment principles.

\vspace{-1pt}
\section{Conclusion}
\label{sec:conclusion}
\vspace{-1pt}

This paper presents CMR-EXTR, a model developed to extract structured data from CMR reports for research data curation and clinical software development. The training pipeline combines knowledge distillation from a large teacher model to a compact student model with human-in-the-loop quality control, effectively reducing annotation effort and enabling efficient offline inference. To improve reliability, we introduce three quality assessment principles that allow the model to generate interpretable confidence estimates. Experimental results demonstrate that CMR-EXTR achieves 99.65\% accuracy in data extraction and 97.04\% accuracy in report classification. These results underscore the potential of large language model-based frameworks for accurate, low-effort extraction of clinically relevant information and highlight their adaptability to a broad range of medical text analysis tasks.

\section{Compliance with ethical standards}
\label{sec:ethics}


This retrospective study used de-identified CMR reports collected previously. The protocol was reviewed and approved by the Institutional Review Board with a waiver of informed consent, given minimal risk and no direct subject contact.

\section{Acknowledgments}
\label{sec:acknowledgments}
This work was supported in part by the National Institutes of Health (R01 HL148103). The authors declare no competing interests.

    
    

\bibliographystyle{IEEEbib}
\bibliography{refs}

\end{document}